%% file: main.tex
\documentclass[letterpaper, 10 pt, conference]{ieeeconf}  

\IEEEoverridecommandlockouts                              

\usepackage{amsmath,amssymb,amsfonts}
\usepackage{dsfont}
\usepackage{bm}
\usepackage{url}
\usepackage{graphics}
\usepackage{times}
\usepackage{algorithmic}
\usepackage{multirow}
\usepackage{booktabs}
\usepackage{multirow}
\usepackage[ruled,linesnumbered]{algorithm2e}
\usepackage{textcomp}
\usepackage{tabularx}
\usepackage{relsize}
\usepackage[flushleft]{threeparttable}
\usepackage{cite}
\newif\ifshortbibnames
\shortbibnamestrue

\usepackage[table,dvipsnames]{xcolor}
\usepackage{colortbl}
\usepackage{enumerate}
\let \labelindent \relax 
\usepackage[inline]{enumitem}
\usepackage{acronym}
\usepackage{siunitx}
\usepackage{placeins}    
\usepackage{subfig}
\usepackage{comment}
\usepackage[font=small]{caption}
\usepackage{wrapfig}
\usepackage{graphicx}
\usepackage{subfig}
\usepackage{lipsum}
\usepackage{tikz}
\usetikzlibrary{positioning, calc, fit, arrows.meta, backgrounds, decorations.pathreplacing, shapes.geometric}
\usepackage{standalone}
\usepackage{amsmath, amssymb}

\usepackage{tikzducks}

\usepackage[most]{tcolorbox}
\definecolor{pastelblue}{RGB}{210,225,245}
\definecolor{pastelblueborder}{RGB}{120,160,210}
\newtcolorbox{resultbox}[1][]{%
    enhanced,
    breakable,
    colback=pastelblue,
    colframe=pastelblueborder,
    arc=4pt,
    outer arc=4pt,
    boxrule=0pt,
    left=6pt, right=6pt, top=6pt, bottom=6pt,
    before skip=6pt, after skip=6pt,
    fonttitle=\bfseries,
    title={Result},
    #1
}

\usepackage[colorinlistoftodos,prependcaption]{todonotes}
\setuptodonotes{inline}

\input{macros}

\makeatletter
\let\NAT@parse\undefined
\renewcommand{\thefigure}{\arabic{section}.\arabic{figure}}
\makeatother
\usepackage[hidelinks]{hyperref}
\usepackage[nameinlink]{cleveref}
\Crefname{figure}{Fig.}{Figs.}
\Crefname{table}{Tab.}{Tabs.}
\Crefname{algorithm}{Alg.}{Algs.}
\Crefname{equation}{Eq.}{Eqs.}

\makeatletter
\renewcommand{\thefigure}{\arabic{figure}}
\makeatother

\title{\LARGE \bf
Residual Compliant Policy: \\ Learning Compliant Policies from Few Demonstrations}

\author{
    Chunyu Jiang$^{12*}$, Yichen Cai$^{2*}$, Shuai Li$^{2}$, Wenbin Sun$^{3}$, Jo\~ao Carvalho$^{24}$,\\Kostas Daniilidis$^{1}$, Oleg Arenz$^{2}$, Cristiana de Farias$^{2}$, Jan Peters$^{2456}$%
    \thanks{
    *Equal contribution.
    $^{1}$GRASP Lab, University of Pennsylvania, PA 19104, USA.
    $^{2}$Intelligent Autonomous Systems Lab, Technical University of Darmstadt, Germany.
    $^{3}$Karlsruhe Institute of Technology, Germany.
    $^{4}$German Research Center for AI (DFKI), SAIROL, Germany.
    $^{5}$Hessian.AI, Germany.
    $^{6}$Robotics Institute Germany (RIG).
    \newline 
    \hspace*{1em} 
    Email: cyjiang@seas.upenn.edu, yichen.cai@tu-darmstadt.de
    }
}

\begin{document}\sloppy


\maketitle
\thispagestyle{empty}
\pagestyle{empty}
\input{sections/0_abstract}

\input{sections/1_introduction}
\input{sections/2_related_works}
\input{sections/3_methodology}
\input{sections/4_experiments}
\input{sections/5_conclusion}

\section*{ACKNOWLEDGMENT}

We thank Nadia Figueroa for sharing valuable insights and Paul Jansonnie for assistance with the teleoperation system.
We acknowledge the use of ChatGPT and Claude to assist in creating the conceptual figures presented in this paper.
This work was funded by the German Research Foundation (DFG) - Project number PE 2315/18-1,  the EU’s Horizon Europe project ARISE - Grant number 101135959, and partially supported by the German Federal Ministry of Research, Technology and Space (BMFTR) under the Robotics Institute Germany (RIG).
This work was also supported by a hardware donation from NVIDIA through the Academic Grant Program, and by the Lichtenberg high-performance computer of TU Darmstadt. 

\bibliographystyle{IEEEtran}
\bibliography{references}






\end{document}

%% file: macros.tex
\definecolor{softred}{RGB}{220,100,80}
\definecolor{softpurple}{RGB}{150,80,190}
\definecolor{softmagenta}{RGB}{190,70,170}
\definecolor{magenta}{RGB}{220,40,190}

%% file: sections/0_abstract.tex
\begin{abstract}
Contact-rich manipulation requires a robot to be precise in free space to
establish the intended contacts, and compliant once in contact to avoid
excessive forces. Imitation learning has proven effective for simpler
manipulation tasks, but the compliance underlying a demonstration is not
directly observable and is therefore difficult to infer from demonstration
data alone. 
We present a framework that makes compliance explicit at every stage.
During data collection, the operator teleoperates the real robot while a
synchronized digital twin visualizes the error between the commanded
reference and the achieved pose, making the current compliance visible,
and selects between two discrete compliance modes. To generalize across
object poses, we combine Laplacian editing, which retargets trajectories
efficiently but is not physics-consistent during contact, with
simulation-based trajectory optimization, which enforces physical
plausibility but offers no manual guidance. Finally, a residual
reinforcement learning policy refines the interaction forces by adjusting the reference pose of the underlying admittance controller.
We evaluate on three flipping and pushing tasks on a Franka FR3 based on a total of 7 demonstrations across tasks. Our method is successful on all 40 out-of-distribution trials, compared to
33/40 and 24/40 for two baselines, with our base policy alone reaching 35/40.
Project website: 
\href{https://residual-compliant-policy.github.io/}{\textcolor{blue}{residual-compliant-policy.github.io}}.
\end{abstract}

%% file: sections/1_introduction.tex
\section{Introduction}
\label{sec:introduction}

\begin{figure}[h]
    \centering
    \includegraphics[width=0.98\linewidth]{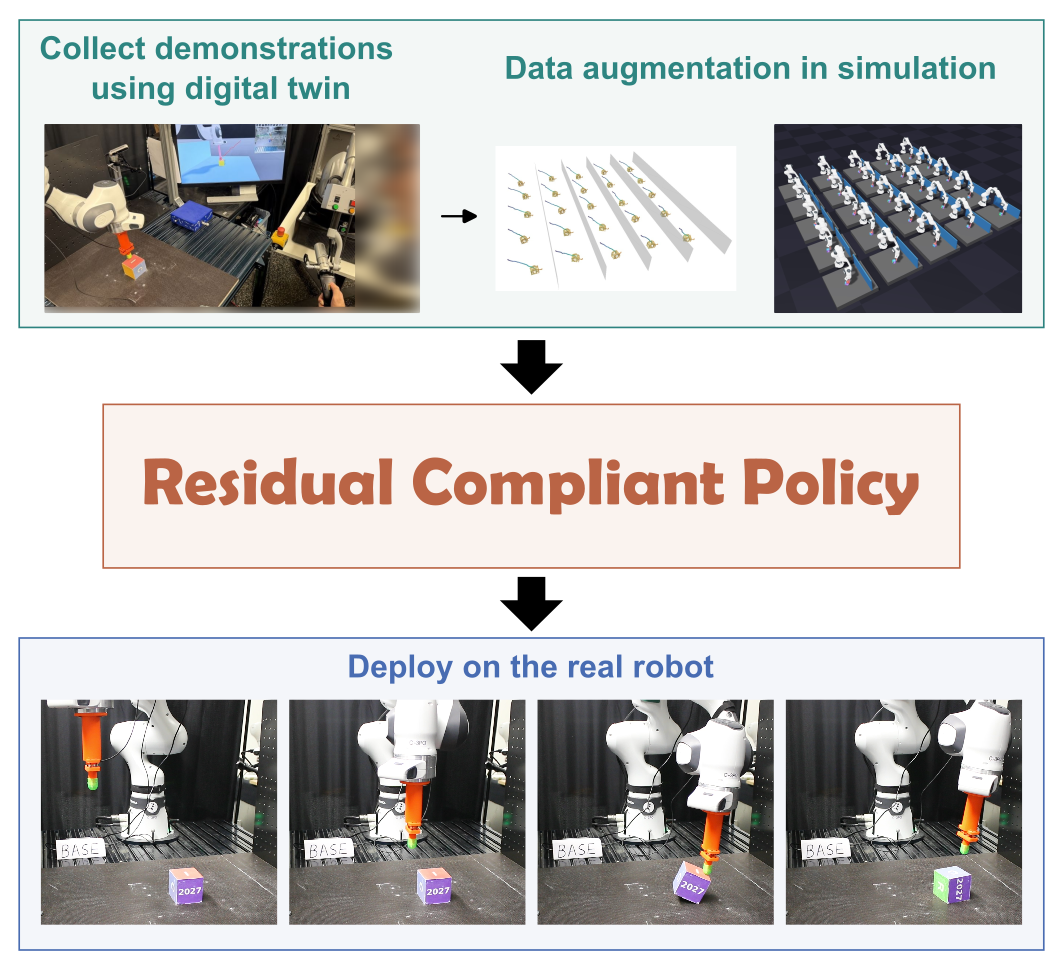}
    \caption{
    {We propose a framework for learning compliant policies from few demonstrations. Trajectories are collected in the real world using a digital twin and subsequently augmented in simulation to train our base policy. We then apply residual RL in simulation and deploy the residual policy in the real world.}}
    \label{fig:placeholder}
    \label{fig:hero}
\end{figure}

Compliance---the ability to yield to external forces---is crucial for robotic manipulation~\cite{whitney1977force, mason1981compliance, hogan1984impedance}, yet the required degree of compliance varies across task stages.
Consider, for instance, the task of flipping a cube with a single finger, as shown in \Cref{fig:hero}. To flip the cube successfully, the robot must first position its fingertip precisely near the cube's edge to establish the intended contact. 
Once the cube begins to tip, the fingertip must follow a suitable motion while maintaining enough force to complete the flip. 
High stiffness helps the fingertip reach the target position accurately.
However, if stiffness is not reduced as contact is established, even small positioning errors can generate large contact forces that may damage the robot.
Conversely, excessively low stiffness may prevent the robot from exerting sufficient force to complete the flip.
Successful execution therefore depends on adapting both motion and compliance throughout the interaction. 
Learning when and how to adapt them, however, is challenging.

Previous work has explored several ways to teach robots compliant behaviors. Reinforcement learning (RL) methods allow robots to acquire these behaviors through interaction~\cite{kalakrishnan2011learning, buchli2011learning, martin2019variable}, but can require extensive trial and error. Collecting this experience on physical robots is costly and risks damaging the robot or its surroundings. Simulation offers a practical alternative, although discrepancies between simulated and real contacts can lead to behaviors that do not transfer reliably. Addressing these discrepancies only after training can require further calibration and policy adaptation, motivating the use of real interaction data to inform simulation from the start.
Imitation learning (IL) thus becomes a natural starting point for learning compliant tasks, as it allows a person to demonstrate how the task should be performed on the real robot~\cite{calinon2010learning, kormushev2011imitation, rozo2016learning}. Demonstrations here can guide both the desired motion and the interaction with the environment, reducing the need to discover the behavior through trial and error~\cite{abu2018force, zeng2021generalization}. However, a demonstrated trajectory does not specify how stiffly the robot should track it or when it should add more compliance. Learning compliant behavior therefore requires additional information about the intended response to contact~\cite{kronander2013learning}. Intuitively conveying this information adds to the demonstration process, which is already time-consuming when diverse task configurations must be covered~\cite{mandlekar2023mimicgen}.

Data augmentation offers a way to reduce this burden by expanding a small set of human demonstrations into a richer training dataset. 
For example, \cite{li2025flow} combines Laplacian editing (LE)~\cite{nierhoff2016spatial} with force modulation~\cite{amanhoud2019dynamical} to generate contact-rich trajectories from a single demonstration. 
LE provides an efficient way to retarget a demonstrated motion to new object poses while preserving its overall shape.
Force modulation encourages sustained contact by offsetting the reference trajectory along the transformed demonstration force~\cite{li2025flow}. 
However, tasks such as aligning a box against a wall also depend on where and in which direction the box is pushed relative to the wall.
These environmental constraints are not explicitly incorporated into object-centric retargeting, so adding a force-informed offset alone may be insufficient to correct the box's orientation.
Trajectory optimization can account for these interactions, while the demonstration provides guidance on how the task should be completed. 
Combining it with LE therefore offers a way to preserve the intent of a demonstration while adapting it to the physical constraints of a new configuration.

In this work, we bring these elements together in a real2sim2real framework for learning compliant non-prehensile manipulation. We begin with demonstrations collected on the real robot, assisted by a synchronized digital twin. During teleoperation, the twin visualizes the difference between the commanded reference and the achieved robot pose, helping the operator see how the robot yields during contact and select between two compliance modes. This makes the intended compliance an explicit part of the demonstration alongside the motion. We then use LE to retarget the demonstrations to new object poses and trajectory optimization with the cross-entropy method (CEM)~\cite{de2005tutorial} to refine the resulting interactions in simulation. From these augmented demonstrations, we learn a state-based compliance policy conditioned on object poses, robot states, and force feedback. These observations are available in simulation and obtained through perception and sensing on the real robot, providing a shared representation for data generation, policy learning, and real-world execution.

Even with data augmentation, the learned policy may encounter contact conditions that are not adequately represented in its demonstrations. Hybrid approaches address this limitation by combining demonstration-based policies with RL refinement~\cite{davchev2022residual, yuan2025policy}. Following this principle, we leverage our simulation pipeline by using residual RL to improve the interaction behavior through corrective actions learned in simulation~\cite{johannink2019residual, davchev2022residual}. The residual policy adjusts the reference pose of the underlying admittance controller, refining the interaction forces while retaining the task behavior acquired from demonstrations. In this way, real demonstrations guide the initial behavior, simulation expands the training data, and residual learning refines execution beyond imitation alone.
We evaluate the framework on three contact-rich manipulation tasks on a Franka FR3 (\Cref{fig:tasks}). Across these tasks, we investigate whether a small set of real demonstrations, augmented and refined in simulation, can support reliable execution under new object configurations.

\begin{itemize}
    \item We introduce a digital twin to collect human demonstrations for contact-rich tasks, providing synchronized visualization of teleoperation commands, object states, and robot states as well as force feedback during teleoperation.

    \item We propose a state-based real2sim2real framework that combines real-world demonstration collection with simulation-based data augmentation, enabling compliance policies learned from simulation to transfer successfully back to the physical robot.

    \item We develop a residual reinforcement learning approach that refines the actions of the learned compliance policy to improve task execution.
\end{itemize}

\begin{figure*}[h]
    \centering
    \includegraphics[width=0.98\linewidth]{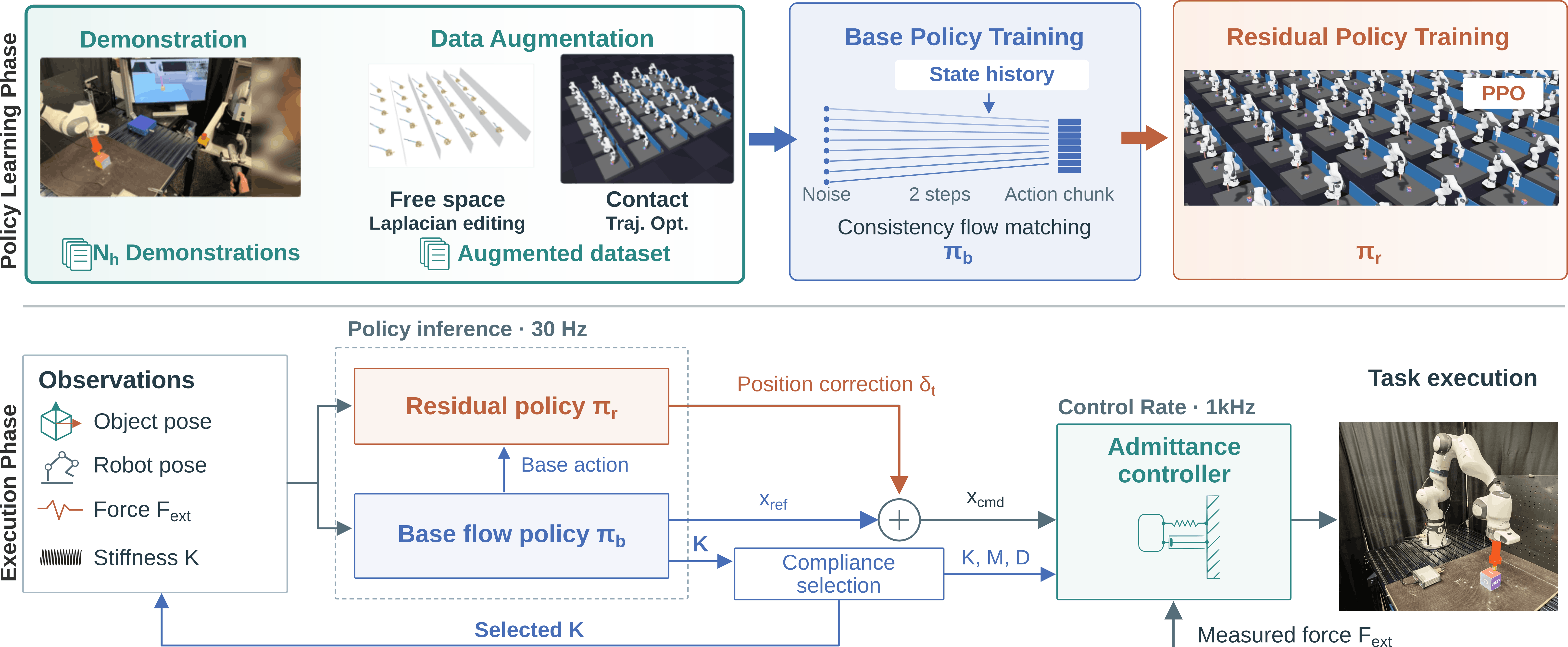}
    \caption{
Proposed compliance policy learning pipeline. 
(a) The operator uses the haptic teleoperation device with our digital twin to collect $N_h$ human demonstrations. 
(b) The collected human demonstrations are augmented via free-space motion Laplacian editing and in-contact trajectory optimization given randomly sampled positions and poses of the object in simulation.
(c) A consistency flow matching-based policy is trained on the augmented dataset.
(d) A residual RL policy is trained upon the base policy in simulation.
(e) The system takes in the robot's proprioception, stiffness of the admittance controller, external force feedback, and the object pose estimated by the digital twin to predict expected stiffness level and reference poses for implementing the task.
    }
    \label{fig:placeholder}
\end{figure*}

%% file: sections/2_related_works.tex
\section{Related Works}
\subsection{Compliant Policy for Contact-Rich Manipulation}
Early work on compliant control primarily focused on developing control paradigms such as impedance control \cite{hogan1984impedance, maples1986experiments}, admittance control \cite{whitney1977force, newman1992stability}, and hybrid force/position control \cite{mason1981compliance, khatib1987unified, nakanishi2008operational} to regulate physical interaction with the environment.
However, these works typically rely on careful design of motion and control parameters.
Some researchers leverage reinforcement learning (RL) \cite{kalakrishnan2011learning, buchli2011learning, martin2019variable} to learn compliance policies without modeling from trial and error.
However, RL policies are typically data-inefficient.
Moreover, contact-rich manipulation requires learning not only the underlying contact dynamics, but also how to approach the target and adapt compliance during interaction, substantially increasing the difficulty of exploration and policy learning.

In contrast, imitation learning (IL) \cite{calinon2010learning, kormushev2011imitation, rozo2016learning, abu2018force, zeng2021generalization, kronander2012online, kronander2013learning, hou2025adaptive, li2025flow, kamijo2024learning, aburub2026learning} can efficiently acquire such skills by leveraging expert demonstrations.
Early work either used Gaussian mixture model (GMM) with Gaussian mixture regression (GMR) \cite{abu2018force, rozo2016learning, kronander2013learning}, or Dynamic Movement Primitives (DMPs) \cite{zeng2021generalization} to learn the robot motion and compliance behaviors.
Recent advances in generative modeling~\cite{ho2020denoising,lipman2022flow,yang2024consistency} have enabled visuomotor policies with force feedback to learn diverse compliance behaviors from multimodal observations~\cite{hou2025adaptive,li2025flow}.
Similarly, we adopt Flow Policy \cite{zhang2025flowpolicy}, a flow-matching variant that leverages consistency flow~\cite{yang2024consistency} for efficient sampling. 
Unlike~\cite{hou2025adaptive,li2025flow}, which use RGB images or point clouds, we use object poses as observations, helping bridge the gap between simulation and the real world \cite{mandlekar2023mimicgen, lin2024twisting}.
Despite efficiently learning compliant behaviors, IL may struggle with disturbances and conditions beyond the demonstrations.
Thus, we propose to learn a residual policy upon the trained IL policy to deal with these problems.

\subsection{Collecting Human Demonstrations for Contact-rich Tasks}
Human demonstrations for contact-rich tasks are commonly collected through kinesthetic teaching~\cite{calinon2010learning, kormushev2011imitation, rozo2016learning, abu2018force, zeng2021generalization, kronander2012online, kronander2013learning, hou2025adaptive} or teleoperation~\cite{kamijo2024learning, aburub2026learning}.
Kinesthetic teaching enables direct physical guidance of the robot, but motion and force recordings alone do not explicitly specify the desired stiffness profile.
Additional estimation is therefore required, using motion variability across demonstrations~\cite{calinon2010learning, kormushev2011imitation} or optimizing stiffness parameters to fit the demonstrated dynamics~\cite{rozo2016learning, abu2018force}.
Alternatively, ~\cite{kronander2012online, kronander2013learning} directly modified the robot stiffness by wiggling and tapping the robot.
However, these perturbations are not applicable during the in-contact phase.

Teleoperation provides a separate interface for commanding robot motion, with force-feedback devices conveying interaction forces.
The closest to ours are ~\cite{kamijo2024learning, aburub2026learning}, which proposed using a VR controller to send the reference position for the robot to follow and a button to set the stiffness level.
However, these interfaces do not explicitly visualize the commanded reference pose within the task scene, whereas our digital twin enables operators to visualize it during teleoperation.
Furthermore, they require collecting 40-60 demonstrations per task, which is time-consuming and labor-intensive.
In \cite{li2025flow}, Laplacian Editing \cite{nierhoff2016spatial} and force modulation \cite{amanhoud2019dynamical} are used to generate new trajectories from a single demonstration collected in simulation.
Building on this idea, we combine Laplacian editing with trajectory optimization \cite{yang2025physics} to efficiently generate high-quality demonstrations.

\subsection{Residual Policy for Contact-Rich Manipulation}
Residual policy learning augments a base policy with a learned correction~\cite{silver2018residual,johannink2019residual}.
Residual corrections can be achieved through online optimization~\cite{zhang2023efficient}, interactive imitation learning from human corrections~\cite{xu2026compliant}, or reinforcement learning~\cite{johannink2019residual, ranjbar2021residual, davchev2022residual, yuan2025policy, ankile2025imitation}.
Most residual RL methods build on pre-designed control laws~\cite{johannink2019residual,ranjbar2021residual} or on demonstrated trajectories represented by probabilistic movement primitives (ProMP) \cite{carvalho2022adapting} or DMPs~\cite{davchev2022residual}.
\cite{yuan2025policy, ankile2025imitation} added residual RL on top of a learned base policy via behavior cloning.
In this work, we explore residual RL on top of a flow-matching-based policy that learns both motion and compliance from demonstrations.

\label{sec:relatedworks}

%% file: sections/3_methodology.tex
\section{Method}
This section covers the main components of our method. 
We first discuss our admittance controller design in \cref{sec:lowlevel}.
Next, we introduce our proposed data collection pipeline in \cref{sec:demo} and data augmentation method in \cref{sec:augmentation}, which generate hundreds of high-quality trajectories from a few collected demonstrations to train the base policy.
We then describe our base policy in \cref{sec:base_policy}.
Finally, we present our residual RL policy in \cref{sec:residual}.

\subsection{Admittance Controller Design}
\label{sec:lowlevel}
To learn a robust compliant manipulation policy from a few demonstrations, we use
a Cartesian space admittance controller \cite{newman1992stability} as the low-level controller, which controls the end-effector to behave like a spring-mass-damper system:
\begin{equation}
\label{eqa:admittance}
\bm{M}\ddot{\bm{x}}
+ \bm{D}\dot{\bm{x}}
+ \bm{K}\left(\bm{x}-\bm{x}_{ref}\right)
= \bm{F}_{ext},
\end{equation}
where $\bm{x}, \dot{\bm{x}}, \ddot{\bm{x}} \in \mathbb{R}^{6}$ denote the Cartesian pose, velocity, and acceleration of the robot's end-effector, respectively.  
$\bm{x}_{ref}$ denotes the reference pose and
$\bm{F}_{ext} \in \mathbb{R}^{6}$ denotes the external wrench exerted on the end-effector. 
$\bm{M}, \bm{D}, \bm{K} \in \mathbb{R}^{6\times6}$ are positive-definite desired inertia, damping, and stiffness matrices, respectively.
We choose each of these matrices to be diagonal, where the coefficients of the inertia matrix are given by ${\bm{M}=\operatorname{diag}(2.0, 2.0, 2.0, 0.5, 0.5, 0.5)}$ and the damping matrix 
 $\bm{D}=\operatorname{diag}(d_1,\ldots,d_6)$ uses critical damping
$d_i = 2\sqrt{m_i k_i}$.
For the stiffness matrix,
$\bm{K}=\operatorname{diag}(k,\ldots,k)$, we deviate from prior work~\cite{hou2025adaptive} by using a single parameter
that changes between $k_h$ during free-space motion and $k_l$ during contact (determined during data collection).
We use two discrete stiffness modes rather than adapting the stiffness
continuously at run time~\cite{hou2025adaptive}, because transitions between a fixed set of modes
can be scheduled to satisfy an explicit stability condition. 
Namely, we interpolate between the two modes along a predefined spline with restricted velocity that strictly satisfies the condition of~\cite{kronander2016stability}. 
Furthermore, the discrete choice keeps the action space small, since the policy only needs to shape the reference trajectory and select the mode. 
As the environments we consider do not exhibit the time-varying
compliance of a human partner \cite{duchaine2007general, ficuciello2015variable}, continuous stiffness adaptation is not necessary.

\subsection{Demonstration Collection}
\label{sec:demo}
\begin{figure}
    \centering
    \includegraphics[width=0.9\linewidth]{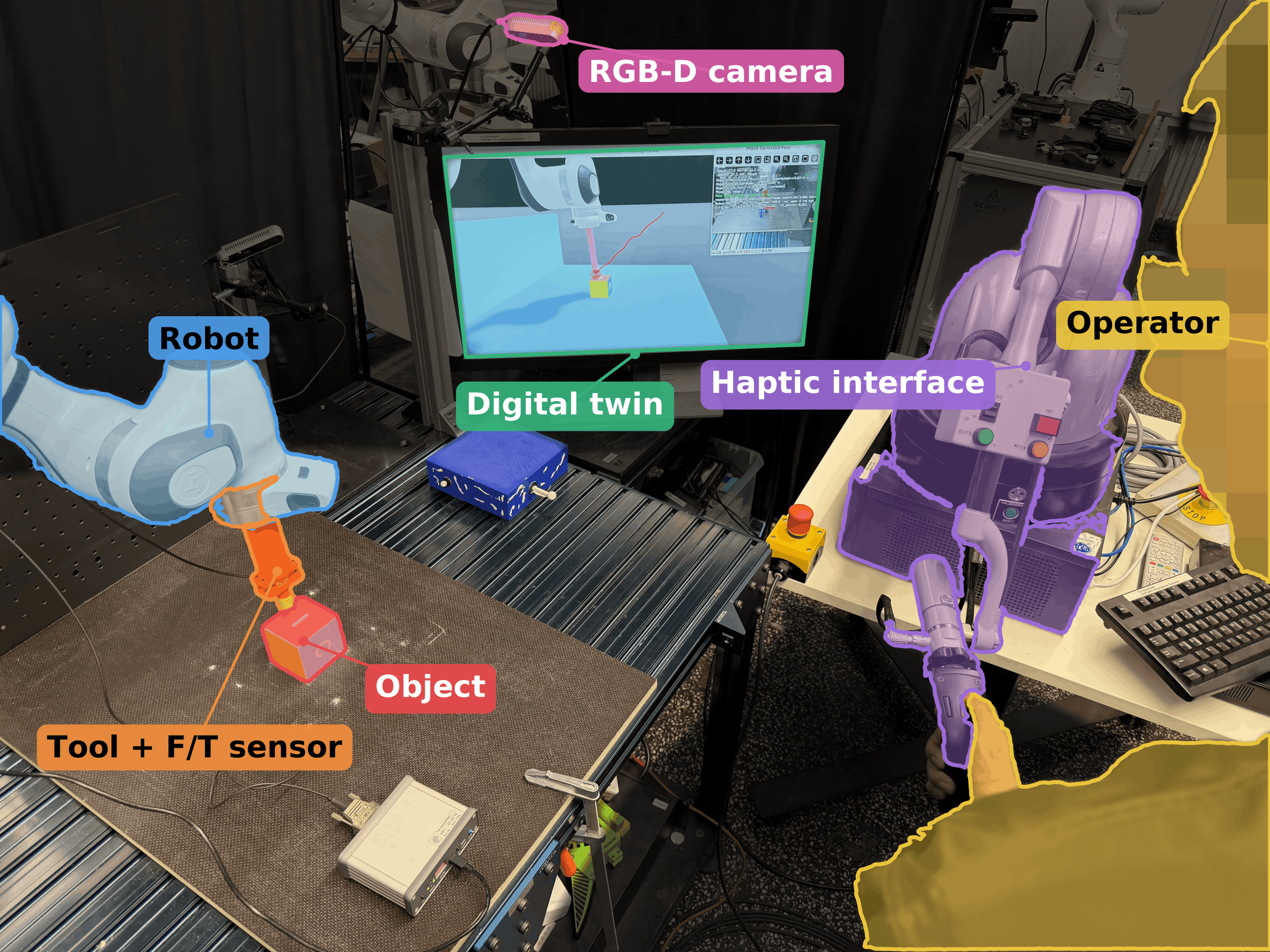}
    \caption{Teleoperation setup. The digital twin synchronizes with the real robot in real-time. }
    \label{fig:teleoperation}
\end{figure}

We collect human demonstrations via teleoperation.
As depicted in \Cref{fig:teleoperation}, our data collection pipeline consists of a Haption Virtuose 6D Tao haptic teleoperation device, a Franka FR3 robot equipped with a force-torque sensor, a RealSense D435 camera, and a Newton \cite{newton2025} simulator, which serves as a digital twin to assist humans with teleoperation.
Our digital twin dynamically synchronizes with the real robot's state.
Specifically, the simulator uses the same admittance controller in \Cref{eqa:admittance} as on the real robot.
The operator moves the haptic device to provide the reference trajectory for the admittance controller, which is sent simultaneously to the simulator and real robot.
For an admittance controller, once the robot makes contact with an object, the reference trajectory and the robot no longer coincide.
Under these circumstances, it is difficult to determine the magnitude and direction of the force applied by observing the real robot alone.
With the visualized reference trajectory in the simulator and the real-time force feedback from the haptic device, the operator can easily feel the force the robot exerts on the object and predict where the robot tends to reach.

Maintaining the twin requires tracking the object's pose, which must
remain accurate under occlusion by the robot and the tool. Hence, we use physics-based prediction with visual correction~\cite{abou2024physically, cai2026gausstwin}.
Since all tasks involve rigid objects, the MuJoCo \cite{todorov2012mujoco} solver is used to predict the object's state given the robot control input.
For visual correction, we adopt FoundationPose \cite{wen2024foundationpose}, which outputs a predicted delta pose given the object's mesh, the object's current pose, and an RGB-D image.
Similar to \cite{abou2024physically, cai2026gausstwin}, a correction velocity proportional to the delta correction pose is added to the current velocity of the object in simulation if the delta correction pose is greater than a preset threshold.
This design offers the following advantages: 
(i) the evolution of objects within the simulation is continuous and satisfies physical constraints;
(ii) FoundationPose pose optimization requires the input object pose to be close to the ground truth, a requirement that can be met via physical predictions in simulation;
(iii) when the object is occluded in the camera view, simulation can maintain tracking and prevent loss of track.
To initialize the digital twin, we first use SAM3 \cite{carion2026sam} to obtain the object's segmentation mask, then run FoundationPose to initialize the object's pose, which is also used to initialize the simulation.
After initialization, our digital twin runs stably at 30~Hz, with the simulation prediction taking 15~ms, the visual correction 10~ms, and other I/O 5~ms.

To collect data, the operator first tunes the low-stiffness value \(k_l\) using a button on the haptic device until the task can be executed comfortably. 
After fixing \(k_l\), the operator moves the haptic device to provide the reference trajectory and uses the button to switch between the two stiffness modes.
Because the robot directly reuses the stiffness selected during demonstration, no force-based stiffness recovery is required, which would otherwise be necessary for demonstrations collected through kinesthetic teaching.
The collected demonstration dataset is denoted as
$
\mathcal{D}_h
=
\left\{
\left(
\bm{\tau}_1^j,
\ldots,
\bm{\tau}_{M_j}^j
\right)
\right\}_{j=1}^{N_h},
$
where $N_h$ is the number of demonstrations and $M_j$ is the number of timestamps of the $j$-th data point.
At each time step $t$, we record
$
\bm{\tau}_t^j = 
\left(
\bm{T}_{R}^{W},
\bm{T}_{O}^{W},
\bm{x}_{ref},
k
\right),
$
where $\bm{T}_{R}^{W}, \bm{T}_{O}^{W} \in SE(3)$ denote the robot end-effector pose and object pose expressed in the world frame, respectively.

\subsection{Demonstration Augmentation}
\label{sec:augmentation}
Starting from $N_h$ human demonstrations, we generate additional demonstrations to expand the training dataset.
Each demonstration is partitioned into free-space and in-contact segments based on its recorded stiffness values.
For free-space motion, we follow~\cite{li2025flow} and apply Laplacian trajectory editing~\cite{nierhoff2016spatial} to adapt the demonstrated end-effector path to new initial and contact positions while approximately preserving its local shape.
Orientations are interpolated separately using spherical linear interpolation (SLERP).
For the in-contact phase, we seek reference trajectories that complete the task while maintaining smooth motion.
We found that force modulation \cite{li2025flow} performs well when the in-contact trajectory is object-centric.
However, for non-object-centric motions, such as nonprehensile box pivoting toward a target pose, force modulation alone is insufficient to generate a suitable trajectory to accomplish the task.
Specifically, we use the cross-entropy method (CEM)~\cite{de2005tutorial} to minimize a cost function subject to the system dynamics, similar to \cite{yang2025physics}.
After optimization, we obtain an augmented dataset
$
\mathcal{D}_a
=
\left\{
\left(
\bm{\tau}_1^j,
\ldots,
\bm{\tau}_{M_j}^j
\right)
\right\}_{j=1}^{N_a},
$
with $N_a$ trajectories to train the base policy.

\subsection{Compliant Base Policy}
\label{sec:base_policy}
Our base policy $\pi_\theta$ receives observations at $30\,\mathrm{Hz}$ and outputs a chunk of reference poses and the stiffness.

\textbf{Policy Input and Output.}
The input of our policy 
$ \bm{o}_t = \left( \bm{x}^{obj}_{t-T:t}, \bm{x}_{t-T:t}, \hat{\bm{F}}_{t-T:t}, k_{t-T:t} \right)$
comprises a four step history ($T=3$) of the object pose $\bm{x}^{obj}_{t-T:t}$, robot end-effector pose $\bm{x}_{t-T:t}$, filtered force feedback from the F/T sensor $ \hat{\bm{F}}_{t-T:t}$, and previously commanded stiffness $k_{t-T:t}$. 
We use a 9D pose representation as in ~\cite{zhou2018continuity}. 
The object poses are provided by our digital twin. 
Compared with RGB-based \cite{hou2025adaptive} and point cloud-based \cite{li2025flow} visual inputs, object poses can be obtained directly in simulation without image rendering or high-dimensional domain randomization, substantially reducing simulation overhead. 
Moreover, state-based policies offer greater training efficiency.
We predict $H=16$ steps, execute $T_a=8$, and keep the remainder as look-ahead. The output $\bm{a}_{t:t+T_a}\in\mathbb{R}^{10}$ consists of a 9D representation of the predicted reference pose $\bm{\hat{x}}_{ref}$ and a predicted stiffness $\hat{k}$.
This simple output makes it easier for the residual policy to modify.

\textbf{Policy Training and Inference.}
We use a multi-segment consistency flow-matching
policy~\cite{yang2024consistency, zhang2025flowpolicy}, realized as a
conditional 1-D U-Net conditioned on the context $\bm{o}_t$ via
FiLM~\cite{perez2017visual}. We use two segments, resulting in two integration steps during inference, which is sufficiently fast to enable replanning within the budget for one policy step. 

Few-step samples carry per-action noise that appears as a staircase in the executed reference, and consecutive chunks, drawn from independent noise, disagree at their join.
We therefore average $N=8$ chunks drawn in a single batched forward pass and cross-fade the first four executed steps with the previous chunk, leaving the stiffness channel unblended.
The resulting goals are fitted per chunk by a clamped quintic B-spline with $C^2$ seams under velocity, acceleration and jerk limits, and streamed as $\bm{x}_{ref}$ at the controller rate.
\vspace{-2pt}
\subsection{Residual Learning}
\label{sec:residual}

The base policy $\pi_\theta$ is trained on demonstrations spanning a limited range of object poses and lacks robustness to dynamics perturbations. Consequently, during deployment, the policy can encounter out of distribution states or new dynamics not seen during demonstrations. This can lead to policy instability and task failure.
Rather than fine-tuning $\pi_\theta$, we freeze it and learn a residual policy $\pi_\phi$ in simulation that corrects its execution in closed loop.
The residual acts on top of the compliant execution stack of \cref{sec:lowlevel}: it perturbs only the reference position tracked by the admittance controller, while $\pi_\theta$ retains authority over the orientation and the stiffness $\hat{k}_t\in\{k_l,k_h\}$.

\textbf{Problem formulation.}
At every step $t$, the base policy provides the current row of its executing chunk (\cref{sec:base_policy}), $\bm{a}_t=(\hat{\bm{x}}_{ref,t},\hat{k}_t)$, where $\hat{\bm{x}}_{ref,t}=(\hat{\bm{p}}_{ref,t},\hat{\bm{R}}_{ref,t})$.
The residual policy $\pi_\phi(\bm{u}_t\mid\bm{o}^{r}_t)$ is conditioned on observation $\bm{o}^{r}_t$, and outputs $\bm{u}_t\in\mathbb{R}^{3}$, which a fixed map turns into a translational offset $\bm{\delta}_t\in\mathbb{R}^{3}$.
The composed command is
\begin{equation}
\label{eqa:compose}
\bm{x}^{cmd}_t=\left(\hat{\bm{p}}_{ref,t}+\bm{\delta}_t,\ \hat{\bm{R}}_{ref,t}\right),
\qquad
k^{cmd}_t=\hat{k}_t ,
\end{equation}
and the execution layer interpolates $\bm{\delta}_t$ linearly between consecutive steps, so the composed command is as smooth as the base's own.

\textbf{Observation Space.}
The residual observes a $T$ step history of an object-centric encoding,
\begin{equation}
\label{eqa:residual_obs}
\bm{o}^{r}_t=\left(\bm{\eta}_{t-T},\ldots,\bm{\eta}_t\right),
\quad
\bm{\eta}_t=\left(
\bm{\xi}_t,\
\bm{a}_t,\
\hat{\bm{F}}_t,\
\bm{e}_t,\
\bm{\delta}_{t-1}
\right),
\end{equation}
where $\bm{\xi}_t$ stacks the object tilt and the end-effector pose relative to the object, $\bm{a}_t$ is the base action, $\hat{\bm{F}}_t$ is the filtered force, $\bm{e}_t=\bm{p}^{cmd}_{t-1}-\bm{p}_t$ is the tracking error, and $\bm{\delta}_{t-1}$ is the previous offset.
Encoding poses relative to the object makes the residual invariant to where in the workspace the task takes place.
Following asymmetric actor--critic~\cite{pinto2017asymmetric}, the critic instead receives $\bm{s}_t=\left(\bm{o}^{r}_t,\bm{z}_t\right)$, where $\bm{z}_t$ holds privileged simulator state: the noise-free wrench, the object twist, its mass and friction coefficients, and the applied perturbation.

\textbf{Reward Functions.}
The reward is sparse in the task and dense only in safety and regularization,
\begin{equation}
\label{eqa:reward}
\begin{aligned}
r_t &= r^{task}_t - r^{safe}_t - r^{reg}_t , \\[2pt]
r^{task}_t &= w_s\,\mathds{1}\!\left[\mathrm{succ}_t\right] - w_f\,\mathds{1}\!\left[\mathrm{fail}_t\right] , \\[2pt]
r^{safe}_t &= w_F\left[\left\lVert\hat{\bm{f}}_t\right\rVert-F_{max}\right]_{+} , \\[2pt]
r^{reg}_t &= w_u\left\lVert\bm{u}_t\right\rVert^{2} + w_\delta \left\lVert\bm{\delta}_t-\bm{\delta}_{t-1}\right\rVert^{2} ,
\end{aligned}
\end{equation}
where $[\cdot]_{+}=\max(\cdot,0)$.
The reward function is composed of the following terms:
1) \textit{Task reward} $r^{task}_t$: a terminal success bonus and a failure penalty.
Success is defined by the task, e.g.\ the object passing its tipping angle and settling.
Failures end the episode without the success bonus: exceeding the contact-force stop or the command-velocity limit, and task-specific conditions under which the goal is reached by an unsafe strategy, such as sliding the object during the flipping.
2) \textit{Force penalty} $r^{safe}_t$: a soft ceiling on the filtered contact force above $F_{max}$, set below the robot's force stop.
Unlike a penalty on contact itself, it leaves the residual free to press harder as long as it stays below the stop.
3) \textit{Regularization penalty} $r^{reg}_t$: penalizes the magnitude of $\bm{u}_t$ and the change of the applied offset, keeping corrections small and smooth so that the composed command stays within the velocity envelope of the base.

\textbf{Training Strategy.}
We optimize $\pi_\phi$ with PPO~\cite{schulman2017proximal} in a Newton simulation~\cite{newton2025} that runs the same admittance controller, execution layer, and safety stops as the real system, with the frozen base policy evaluated inside the loop.
Each episode randomizes the object pose, mass, and friction coefficients, occasionally perturbs the object kinematically before contact, and corrupts the observed object pose and wrench with per-episode biases and per-step noise matched to the perception stack.
The output layer of $\bm{\mu}_\phi$ is initialized to zero, so training starts exactly from the base policy and exploration remains in its success basin.
\label{sec:method}

%% file: sections/4_experiments.tex
\section{Experiments}
In this section, we evaluate our framework through experiments designed to answer the following questions: 
(i) Can our data collection and augmentation pipeline provide sufficient high-quality trajectories for policy learning? 
(ii) Can a policy trained on the generated data successfully perform contact-rich manipulation on the real robot? 
(iii) How does our compliance strategy compare with an ACP-style compliance policy? 
(iv) Does residual reinforcement learning improve the performance of the base policy?

\subsection{Experimental Setup}
We evaluate our method on a Franka FR3 robot with a Resense HEX21 force/torque (F/T) sensor near the pusher tip and an external RealSense D435 camera. 
The pusher and its end cap are both 3D-printed. 
RGB-D images and F/T measurements are streamed at 30 Hz and 1 kHz, respectively.
A haptic Haption Virtuose 6D Tao is employed to teleoperate the robot.

\begin{figure}[t]
    \centering
    \includegraphics[width=0.98\linewidth]{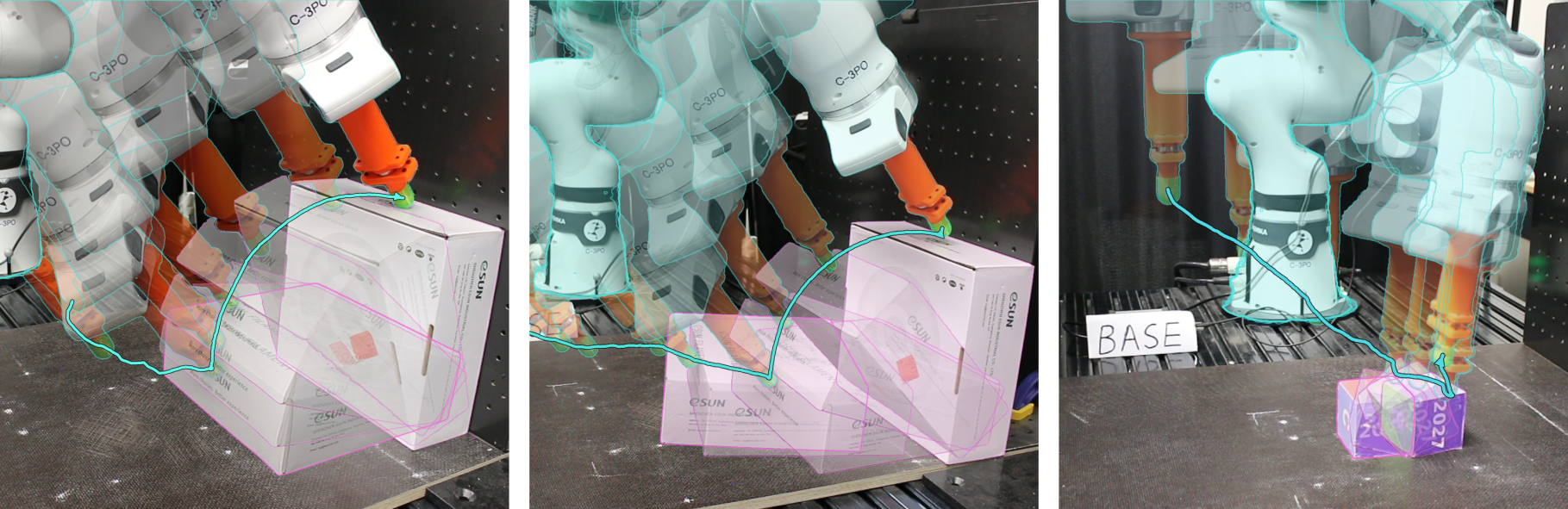}
    \caption{
    {Real-robot experiments on three tasks: \emph{flip box}, \emph{push and flip box}, and \emph{flip cube} (left to right).}
    }
    \label{fig:tasks}
\end{figure}

\textbf{Tasks.}
All policies are evaluated on three tasks as shown in \Cref{fig:tasks}: \emph{flip box}, pivoting a box upright against a wall with a pusher; \emph{push and flip box}, pushing the box against the wall before pivoting it upright; and \emph{flip cube}, pivoting a cube over on the table. 
Each policy is evaluated in 30 in-distribution trials per task. 
For \emph{push and flip box} and \emph{flip cube}, we additionally run 20 hard trials per policy, with five trials under each of four conditions: object positions outside the training range, object rotations outside the training range, an external shove applied to the object during execution, and an unseen object with greater mass and lower friction (Obj.~B in Table~\ref{tb:flipping_result}). 
The nominal box is a 147 g paper box, whereas Box B is a geometrically identical 201 g 3D-printed box. Similarly, the nominal cube is a 100 g wooden cube covered with paper, whereas Cube B is a geometrically identical 131 g 3D-printed cube.

\textbf{Baselines.}
We compare our base policy and residual policy to two baselines: (a)  \textbf{Adaptive Compliance Policy (ACP)} \cite{hou2025adaptive}, which predicts a 19-D action that includes the reference pose, virtual target pose, and a scalar value representing the stiffness magnitude in the low-stiffness direction. 
Here, labels are generated from our augmented dataset to train the base policy to output the same predictions as ACP.
And, 
(b) \textbf{LE},  based on the same Laplacian editing data augmentation method from \cite{li2025flow} to train our base policy.

A single demonstration is collected for \emph{flip box}. 
Three demonstrations are collected for \emph{flip cube} and \emph{push and flip box} with the object initially lying flat at yaw angles of roughly $0^\circ$ and $\pm15^\circ$, respectively, as multi-modality occurs in these two tasks.
The behavior cloning policies are trained for 600 epochs using 150 augmented trajectories for \emph{push and flip box} and \emph{flip cube}, and 400 epochs with 40 trajectories for \emph{flip box}. 
The residual RL policy is trained for 1000 iterations.
All policies run at 30 Hz.

\textbf{Metrics.}
The four policies are evaluated in terms of success rate, average peak contact force, and average completion time.
A trial is considered successful if the object is rotated by more than $85^\circ$.

\vspace{-5pt}
\begin{figure}[h]
    \centering
    \includegraphics[width=1.0\linewidth]{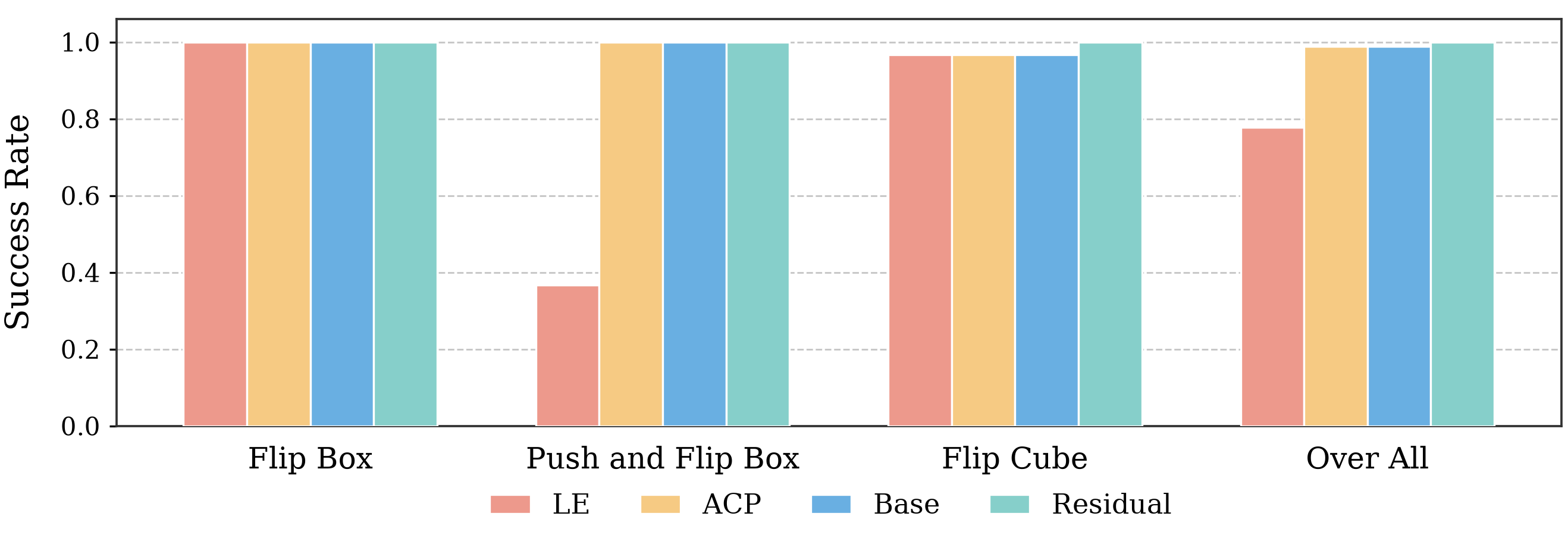}
    \vspace{-15pt}
    \caption{Success rate of in-distribution tests.}
    \label{fig:sr}
\end{figure}
\vspace{-10pt}

\subsection{Results}

\textbf{Policies trained solely on generated trajectories succeed on the real robot (i, ii).}
Fig.~\ref{fig:sr} and Table~\ref{tb:flipping_result} summarize 520 real-robot trials (130 per policy). 
In distribution (30 trials per task), the base policy trained on the 150 augmented trajectories succeeds in 100\% of \emph{flip box} and \emph{push and flip box} trials and 96.7\% of \emph{flip cube} trials (89/90 overall), and the residual policy in all 90. 
\emph{flip box} is solved by every policy, LE included, so the comparisons below rest on the two harder tasks: \emph{push and flip box} adds a pushing phase whose end is fixed by the wall, and \emph{flip cube} requires a precise landing on a small footprint.
\begin{table*}[th]
\centering
\caption{Hard tests success rate (successes/total trials).
Mean pools results from both tasks.}
\label{tb:flipping_result}
\setlength{\tabcolsep}{4pt}
\resizebox{\textwidth}{!}{%
\begin{tabular}{l*{15}{c}}
\toprule
& \multicolumn{5}{c}{Push and Flip Box}
& \multicolumn{5}{c}{Flip Cube}
& \multicolumn{5}{c}{Mean} \\
\cmidrule(lr){2-6}
\cmidrule(lr){7-11}
\cmidrule(lr){12-16}
& Pos & Rot & Shove & Obj B & Overall
& Pos & Rot & Shove & Obj B & Overall
& Pos & Rot & Shove & Obj B & Overall \\
\midrule
LE
& 2/5 & 1/5 & 1/5 & 3/5 & 7/20
& 4/5 & 4/5 & 4/5 & \textbf{5/5} & 17/20
& 6/10 & 5/10 & 5/10 & 8/10 & 24/40 \\

ACP
& 4/5 & 3/5 & \textbf{5/5} & \textbf{5/5} & 17/20
& 4/5 & 3/5 & 4/5 & \textbf{5/5} & 16/20
& 8/10 & 6/10 & 9/10 & \textbf{10/10} & 33/40 \\

Base
& 4/5 & 4/5 & \textbf{5/5} & \textbf{5/5} & 18/20
& 4/5 & 4/5 & \textbf{5/5} & 4/5 & 17/20
& 8/10 & 8/10 & \textbf{10/10} & 9/10 & 35/40 \\

Residual
& \textbf{5/5} & \textbf{5/5} & \textbf{5/5}
& \textbf{5/5} & \textbf{20/20}
& \textbf{5/5} & \textbf{5/5} & \textbf{5/5}
& \textbf{5/5} & \textbf{20/20}
& \textbf{10/10} & \textbf{10/10} & \textbf{10/10}
& \textbf{10/10} & \textbf{40/40} \\
\bottomrule
\end{tabular}%
}
\vspace{-5mm}
\par\vspace{6pt}
\end{table*}

\textbf{Our augmentation outperforms Laplacian editing when multiple objects constrain the motion (i).}
LE and the three policies trained on our data share the backbone, data budget, and schedule, so their gap isolates the augmentation. On \emph{push and flip box}, LE succeeds in 11/30 trials (36.7\%), while ACP, base and residual all reach 100\%. Laplacian editing retargets the seed trajectory to the perturbed object pose but is agnostic to the environment: the box must first be pushed against the wall, and it is the wall, not the initial box pose, that fixes where pushing ends and pivoting begins. A trajectory deformed with respect to the object alone reaches the wall too early or too late, and the flip fails. Our augmentation generates trajectories consistent with both the object and the wall, 
so the contact transition is preserved across perturbations. \emph{flip cube}, in contrast, is object-centric: the cube pivots on the table, the contact constraint moves with the object, and retargeting alone suffices---LE reaches 96.7\%, on par with ACP and base.

\textbf{A two-level stiffness profile with a compliant reference pose matches or exceeds continuous stiffness adaptation (iii).}
ACP and our base policy differ only in the action space: ACP predicts a 19-D action containing the reference pose, a virtual target pose and a continuously varying stiffness magnitude along a predicted low-stiffness direction, whereas our base policy predicts a reference pose and a stiffness level $k\in\{k_l,k_h\}$. In-distribution, the two are indistinguishable (89/90 each). Under perturbations, the base policy succeeds in 35/40 trials against 33/40 for ACP, matching or exceeding it in seven of eight splits, with the largest gap under rotational perturbation (8/10 vs.\ 6/10). Two effects explain this. First, regressing a stiffness magnitude and direction on top of two poses enlarges the target the policy must fit from the same 150 trajectories, and an error in the predicted low-stiffness direction directly corrupts the contact behavior; holding $k=k_l$ during contact and moving the reference pose instead realizes the same compliance through a lower-dimensional, more accurately regressed action. Second, isotropic stiffness is invariant to the orientation of the contact normal, so a rotated object changes nothing in the compliance parameters, whereas ACP must extrapolate its stiffness direction. Continuously adapting compliance parameters is thus not required for these tasks: compliance can be realized by adjusting the reference pose around a fixed two-level stiffness profile.

\begin{figure}[h]
    \centering
    \includegraphics[width=1.0\linewidth]{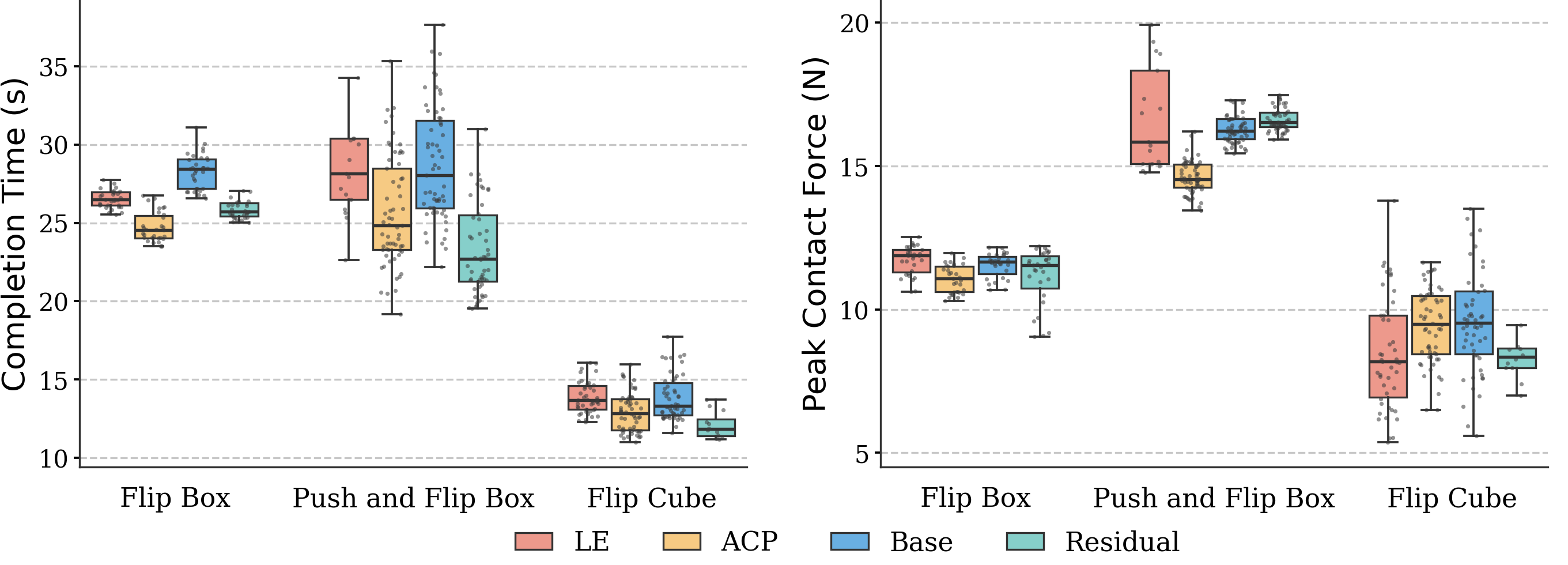}
    \vspace{-15pt}
    \caption{Average completion time (left) and average peak force (right)}
    \label{fig:time4}
\end{figure}

\textbf{Residual RL closes the remaining gap and extends robustness beyond the demonstrations (iv).}
The residual policy is the only one to succeed in every trial, 90/90 in-distribution and 40/40 hard tests, against 89/90 and 35/40 for the base policy it corrects. The gains concentrate where behavior cloning is weak. 
In distribution, the single base failure is on \emph{flip cube}, whose small footprint requires the tool to land precisely; the residual removes it. 
In hard tests, the residual recovers all five base failures: position and rotation perturbations on both tasks (4/5 $\rightarrow$ 5/5 each) and the heavier, more slippery object on \emph{flip cube}. These failures are of a kind imitation cannot fix from a fixed dataset---the demonstrations contain no recovery from a misplaced pivot or a slipping contact---whereas reward-driven residual learning trains on exactly these outcomes and learns corrections to the base action that re-establish and hold contact. The improvement carries over to execution quality (Fig.~\ref{fig:time4}): against the base policy, the residual completes every task 7--18\% faster at comparable or lower peak contact force (10.0 $\rightarrow$ 8.5\,N on \emph{flip cube}, with a markedly tighter spread), so its gains are not bought with more aggressive contact. Mean hard tests success rises monotonically along the pipeline: 60.0\% (LE), 82.5\% (ACP), 87.5\% (base), 100\% (residual).

\vspace{-5pt}

%% file: sections/5_conclusion.tex
\section{Conclusion}
\vspace{-1pt}
We presented a framework for learning compliant manipulation from a handful of demonstrations that leverages an explicitly user-provided binary compliance selection and a residual policy that tunes the contact forces by adapting the position references of the compliant controller. Our method further builds on a digital twin for visualizing compliance during demonstrations on the real robot, and a combination of Laplacian editing and trajectory optimization for efficient and physically consistent data
augmentation.

On three flipping tasks, our method succeeds on all 40 out-of-distribution trials, compared to 33/40 for a state-of-the-art adaptive compliance baseline. Our base policy alone already reaches 35/40 against the baseline's 33/40, and the residual improves this to 40/40.

For future work, the compliance representation could be extended beyond a single scalar stiffness applied to all axes, which suffices for pivoting with a rotationally symmetric tool but may be limiting for tasks requiring direction-dependent compliance, such as insertion. Here, selecting appropriate frames for expressing anisotropic stiffnesses, scheduling transitions between them without losing the stability guarantee, and
providing the user an interface for teaching anisotropic compliance while in contact are the main challenges that would need to be addressed. For tasks that require precise control of moments, such as screwing motions, we further plan to extend the residual policy to adapt the full reference pose.